\def\year{2021}\relax
\documentclass[letterpaper]{article} % DO NOT CHANGE THIS
\usepackage{aaai21}  % DO NOT CHANGE THIS
\usepackage{times}  % DO NOT CHANGE THIS
\usepackage{helvet} % DO NOT CHANGE THIS
\usepackage{courier}  % DO NOT CHANGE THIS
\usepackage[hyphens]{url}  % DO NOT CHANGE THIS
\usepackage{graphicx} % DO NOT CHANGE THIS
\usepackage{natbib}  % DO NOT CHANGE THIS AND DO NOT ADD ANY OPTIONS TO IT
\usepackage{caption} % DO NOT CHANGE THIS AND DO NOT ADD ANY OPTIONS TO IT
\usepackage{amsmath}
\usepackage{amssymb}
\usepackage{dsfont}
\usepackage{booktabs}
\usepackage{multirow}

\usepackage{subcaption}
\title{Discovering New Intents with Deep Aligned Clustering}

\author{
	Hanlei Zhang,\textsuperscript{\rm 1, 2}
	Hua Xu,\textsuperscript{\rm 1, 2}\thanks{Hua Xu is the corresponding author.}
	Ting-En Lin\textsuperscript{\rm 1, 2},
	Rui Lyu\textsuperscript{\rm 1, 3}\\
}

\affiliations{
	\textsuperscript{\rm 1}State Key Laboratory of Intelligent Technology and Systems, \\ 
Department of Computer Science and Technology, Tsinghua University, Beijing 100084, China, \\
\textsuperscript{\rm 2} Beijing National Research Center for Information Science and Technology (BNRist), Beijing 100084, China\\
\textsuperscript{\rm 3} Beijing University of Posts and Telecommunications University, Beijing 100876, China\\
zhang-hl20@mails.tsinghua.edu.cn, xuhua@tsinghua.edu.cn, ting-en.lte@alibaba-inc.com, lvrui2017@bupt.edu.cn

}

\begin{document}
	\maketitle
	%\linenumbers
	\begin{abstract}
		Discovering new intents is a crucial task in dialogue systems. Most existing methods are limited in transferring the prior knowledge from known intents to new intents. They also have difficulties in providing high-quality supervised signals to learn clustering-friendly features for grouping unlabeled intents. In this work, we propose an effective method, Deep Aligned Clustering, to discover new intents with the aid of the limited known intent data. Firstly, we leverage a few labeled known intent samples as prior knowledge to pre-train the model. Then, we perform k-means to produce cluster assignments as pseudo-labels. Moreover, we propose an alignment strategy to tackle the label inconsistency problem during clustering assignments. Finally, we learn the intent representations under the supervision of the aligned pseudo-labels. With an unknown number of new intents, we predict the number of intent categories by eliminating low-confidence intent-wise clusters. Extensive experiments on two benchmark datasets show that our method is more robust and achieves substantial improvements over the state-of-the-art methods. The codes are released at \url{https://github.com/thuiar/DeepAligned-Clustering}.
	\end{abstract}
	
	\section{Introduction}
	Discovering novel user intents is important to improve the service quality in dialogue systems. By analyzing the discovered new intents, we may find underlying user interests, which could provide business opportunities and guide the improvement direction~\cite{lin-xu-2019-deep}.

	Intent discovery has attracted much attention in recent years~\cite{perkins-yang-2019-dialog,ijcai2020-532,10.1145/3366423.3380268}. Many researchers regard it as an unsupervised clustering problem, and they manage to incorporate some weak supervised signals to guide the clustering process. For example,~\citet{hakkani-tr2013a} propose a hierarchical semantic clustering model and collect web page clicked information as implicit supervision for intent discovery.~\citet{hakkani2015clustering} utilize a semantic parsing graph as extra knowledge to mine novel intents during clustering.~\citet{Padmasundari2018} benefit from the consensus predictions of multiple clustering techniques to discover similar semantic intent-wise clusters.~\citet{haponchyk2018supervised} cluster questions into user intent categories under the supervision of structured outputs.~\citet{shi2018auto} extract intent features with an autoencoder and automatically label the intents with a hierarchical clustering method.
	\begin{figure}[t!]
		\centering  
		\includegraphics[scale=.4]{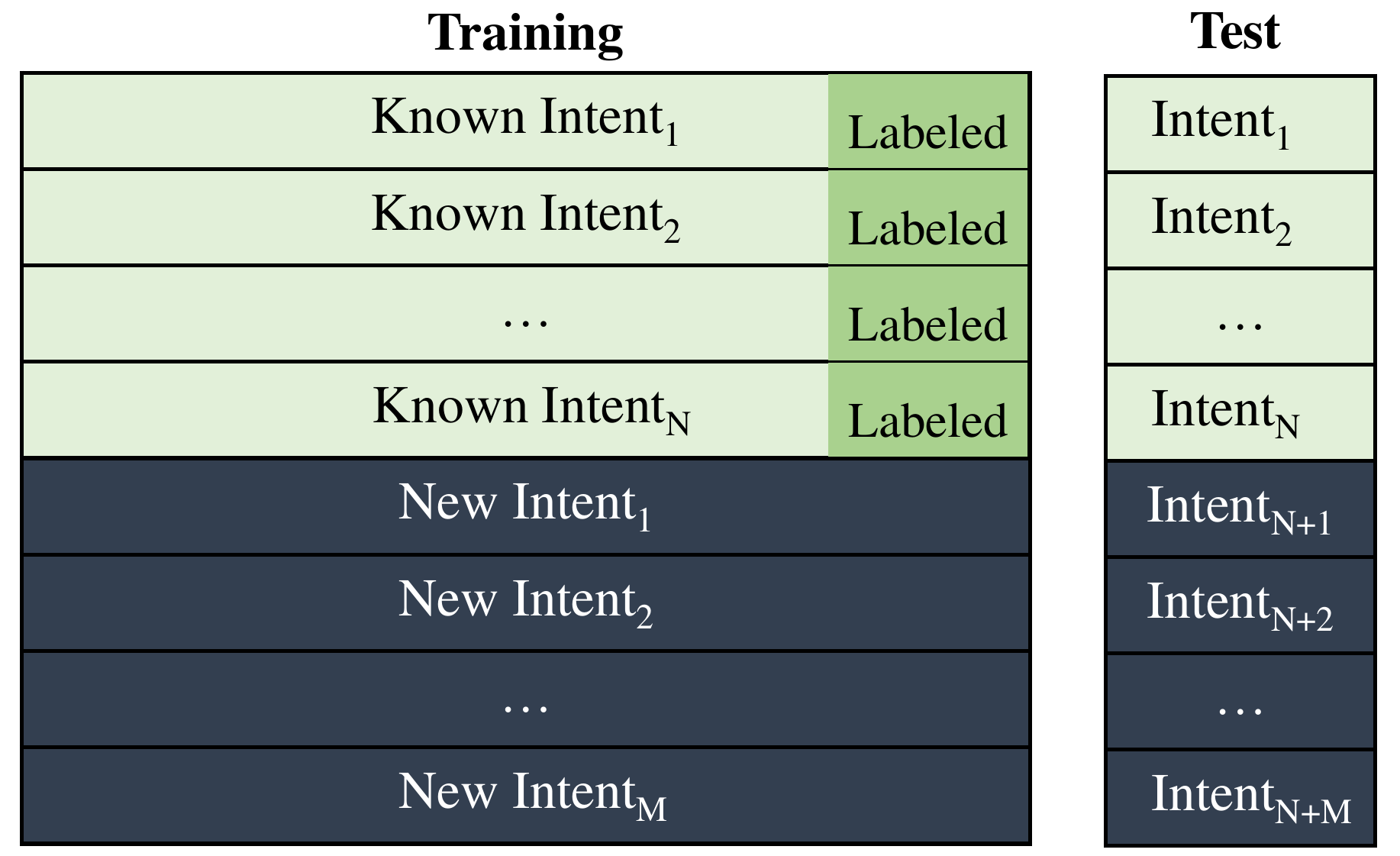}
		\caption{\label{example} An example for our task. We use limited known intent labeled data as a guide to discover new intents. }
	\end{figure}
	\begin{figure*}
		\centering
		\includegraphics[scale=.6]{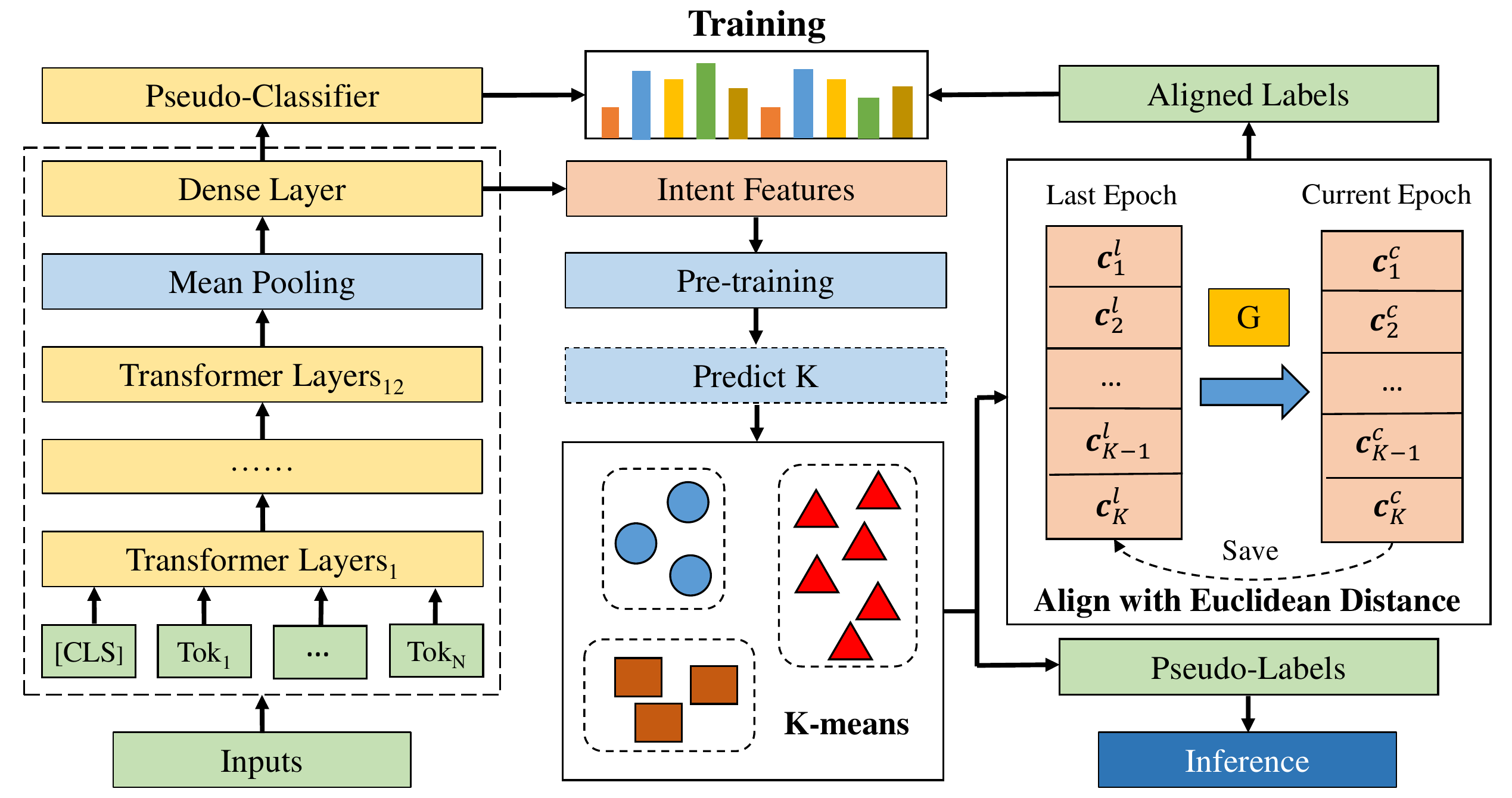}
		\caption{The model architecture of our approach. Firstly, we extract intent features with BERT. We pre-train the model under the supervision of few labeled samples, and predict the cluster number $K$ if we do not know in advance. Then, we perform k-means to produce cluster centroids and use cluster assignments as pseudo-labels. Next, we align the obtained centroids in the current training epoch $\{c_{i}^{c}\}_{i=1}^{K}$ with the saved centroids in the last epoch $\{c_{i}^{l}\}_{i=1}^{K}$, and produce the alignment projection $G$. Finally, we use $G$ on the pseudo-labels to produce the aligned labels for self-supervised learning.}
		\label{model}
	\end{figure*}

	However, all of the above methods fail to leverage the prior knowledge of known intents. These methods assume that the unlabeled samples are only composed of undiscovered new intents. A more common case is that some labeled data of known intents are accessible and the unlabeled data are mixed with both known and new intents. As illustrated in Figure~\ref{example}, we may have a few labeled samples (e.g., with a labeled proportion of 10\%) of known intents in advance. The remaining known and new intent samples are all unlabeled. Our goal is to find known intents and discover new intents with the prior knowledge of limited labeled data. Our previous work CDAC+~\cite{lin2020discovering} directly tackles this problem. Nevertheless, it uses pairwise similarities as weak supervised signals, which are ambiguous to distinguish a mixture of unlabeled known and new intents. Thus, the performance drops with more new intents.
	
	To summarize, there are two main difficulties in our task. On the one hand, it is challenging to effectively transfer the prior knowledge from known intents to new intents with limited labeled data. On the other hand, it is hard to construct high-quality supervised signals to learn friendly representations for clustering both unlabeled known and new intents.
	
	To solve these problems, we propose an effective method to leverage the limited prior knowledge of known intents and provide high-quality supervised signals for feature learning.  As illustrated in Figure~\ref{model}, we firstly use the pre-trained BERT model~\cite{devlin2018bert} to extract deep intent features. Then, we pre-train the model with the limited labeled data under the supervision of the softmax loss. We retain the pre-trained parameters and use the learning information to obtain well-initialized intent representations. Next, we perform clustering on the extracted intent features and estimate the cluster number $K$ (unknown beforehand) by eliminating the low-confidence clusters.
	
	As most of the training samples are unlabeled, we propose an original alignment strategy to construct high-quality pseudo-labels as supervised signals for learning discriminative intent features. For each training epoch, we firstly perform k-means on the extracted intent features, and then use the produced cluster assignments as pseudo-labels for training the neural network. However, the inconsistent assigned labels cannot be directly used as supervised signals, so we use the cluster centroids as the targets to obtain the alignment mapping between pseudo-labels in consequent epochs. Finally, we perform k-means again for inference. Benefit from the relatively consistent aligned targets, our method can inherit the history learning information and boost the clustering performance.
	
	We summarize our contributions as follows. Firstly, we propose a simple and effective method that successfully generalizes to mass of new intents and estimate the number of novel classes with limited prior knowledge of known intents. Secondly, we propose an effective alignment strategy to obtain high-quality self-supervised signals by learning discriminative features to distinguish both known and new intents. Finally, extensive experiments on two benchmark datasets show our approach yields better and more robust results than the state-of-the-art methods. 
	
	\section{Related Work}
	\subsection{Intent Modeling}
	Many researchers try modeling user intents in dialogue systems in recent years. A line for these works is to enrich the intent information jointly with other tasks, such as sentiment classification~\cite{Qin_Che_Li_Ni_Liu_2020}, slot filling~\cite{qin-etal-2019-stack,goo-etal-2018-slot,wang-etal-2018-bi} and so on. Another line is to 
	leverage hidden semantic information to construct supervised signals for intent feature learning~\cite{shi2018auto,Brychcin2017UnsupervisedDA,hakkani-tr2013a}. In this work, we follow the second line to model intents. 
	
	\subsection{Unsupervised Clustering}
	There are many classical unsupervised clustering methods, such as partition-based methods~\cite{macqueen1967some}, hierarchical methods~\cite{gowda1978agglomerative} and density-based methods~\cite{ester1996density}. However, the high-dimensional pattern representations suffer from high computational complexity and poor performance. Though some feature dimensionality reduction~\cite{1984A} and data transformation methods~\cite{wold1987principal} have been proposed, these methods still can not capture high-level semantics of intent features~\cite{lin-xu-2019-deep}. 
	
	\subsubsection{Deep Clustering}
	With the development of deep learning, researchers adopt deep neural networks (DNNs) to extract friendly features for clustering. The joint unsupervised learning (JULE)~\cite{yang2016joint} combines deep feature learning with hierarchical clustering but needs huge computational and memory cost on large-scale datasets. Deep Embedded Clustering (DEC)~\cite{xie2016unsupervised} trains the autoencoder with the reconstruction loss and iteratively refines the cluster centers by optimizing KL-divergence with an auxiliary target distribution. Compared with DEC, Deep Clustering Network (DCN)~\cite{yang2017towards} further introduces a k-means loss as the penalty term to reconstruct the clustering loss. Deep Adaptive Image Clustering (DAC)~\cite{chang2017deep} utilizes the pairwise similarities as the learning targets and adopts an adaptive learning algorithm to select samples for training. However, all these clustering methods cannot provide specific supervised signals for representation learning.
	
	DeepCluster~\cite{caron2018deep} benefits from the structured outputs to boost the discriminative power of the convolutional neural network (CNN). It alternately performs k-means and representation learning. It considers the cluster assignments as pseudo-labels, which are explicit supervised signals for grouping each class. However, it needs to reinitialize the classifier parameters randomly before each training epoch. To deal with this issue, we propose an alignment strategy to produce aligned pseudo-labels for self-supervised learning without reinitialization.
	
	\subsection{Semi-supervised Clustering}
	Although there are various unsupervised clustering methods, the performances of these methods are still limited without the prior knowledge for guiding the clustering process. Therefore, researchers perform semi-supervised clustering with the aid of some labeled data.
	
	Classical constrained clustering methods use the pairwise information as constraints for guiding the representation learning and clustering process. COP-KMeans~\cite{Wagstaff2001} uses instance-level constraints (must-link and cannot-link) and modifies k-means to satisfy these constraints. PCK-means~\cite{basu2004active} presents a framework for pairwise constrained clustering, and it further selects informative pairwise constraints with an active learning method. MPCK-means~\cite{bilenko2004integrating} incorporates the metric-learning approach into PCK-means and combined the centroid-based methods and metric-based methods into a unified framework. However, these methods need huge computational cost by enumerating pairwise conditions. 
	
	KCL~\cite{hsu2018learning} uses deep neural networks to perform pairwise constraint clustering. It firstly trains an extra network for binary similarity classification with a labeled auxiliary dataset. Then, it transfers the prior knowledge of pairwise similarity to the target dataset and uses KL-divergence to evaluate the pairwise distance. MCL~\cite{hsu2018multiclass} uses the meta classification likelihood as the criterion to learn pairwise similarities. However, the domain adaptation methods are still limited in our task. CDAC+~\cite{lin2020discovering} is specifically designed for discovering new intents. It uses limited labeled data as a guide to learn pairwise similarities. However, it is limited in providing specific supervised signals and fails to estimate the number of novel classes. DTC~\cite{Han2019learning} is a method for discovering novel classes in computer vision. It improves the DEC algorithm and transfers the knowledge of labeled data to estimate the number of novel classes. However, the amount of the labeled data has a great influence on its performance.
	
	\begin{table*}[t!]
		\centering
		\begin{tabular}{@{} ccccccc @{}}
			\toprule
			Dataset & \#Classes (Known + Unknown) & \#Training & \#Validation & \#Test & Vocabulary & Length (max / mean) \\
			\midrule
			CLINC & 150 (113 + 37) & 18,000 & 2,250 & 2,250 & 7,283 & 28 / 8.31 \\
			BANKING & 77 (58 + 19) & 9,003 & 1,000 & 3,080 & 5,028 & 79 / 11.91 \\ 
			\bottomrule
		\end{tabular}
		\caption{ \label{datasets}  Statistics of CLINC and BANKING datasets. \# indicates the total number of sentences. In each run of the experiment, we randomly select 75\% intents as known intents. Taking the CLINC dataset as an example, we randomly select 113 known intents and treat the remaining 37 intents as new intents. }
	\end{table*}

	\section{Our Approach}
	In this section, we will describe the proposed method in detail. As shown in Figure~\ref{model}, we firstly extract intent representations with BERT. Then, we transfer the knowledge from known intents with limited labeled data. Finally, we propose an alignment strategy to provide self-supervised signals for learning clustering-friendly representations. 
	
	\subsection{Intent Representation}
	The pre-trained BERT model demonstrates its remarkable effect in NLP tasks~\citep{devlin2018bert}, so we use it to extract deep intent representations. Firstly, we feed the $i^{th}$ input sentence $\boldsymbol{s}_{i}$ to BERT, and take all its token embeddings $[CLS, T_1, \cdots, T_M]$ $\in$ $\mathds R^{(M+1) \times H}$ from the last hidden layer. Then, we apply mean-pooling to get the averaged sentence feature representation $\boldsymbol{z}_{i} \in \mathds R^{H}$:
	\begin{align}
		\boldsymbol{z}_{i} = \text{mean-pooling}([CLS, T_1, \cdots, T_M]),
	\end{align}
	where $CLS$ is the vector for text classification, $M$ is the sequence length, and $H$ is the hidden size. To further enhance the feature extraction capability, we add a dense layer $h$ to get the intent feature representation $\boldsymbol{I}_{i} \in \mathds R^{D}$:
	\begin{align}
		\boldsymbol{I}_{i}=h(\boldsymbol{z}_i) = \sigma(W_h\boldsymbol{z}_{i}+b_h),
	\end{align}
	where $D$ is the dimension of the intent representation, $\sigma$ is the Tanh activation function,  $W_h \in \mathds R^{H \times D}$ is the weight matrix and $b_h \in \mathds R^{D}$ is the corresponding bias term.
	
	\subsection{Transferring Knowledge from Known Intents}
	To effectively transfer the knowledge, we use the limited labeled data to pre-train the model and leverage the well-trained intent features to estimate the number of clusters.
	
	\subsubsection{Pre-training} We hope to incorporate the limited prior knowledge to obtain a good representation initialization for grouping both known and novel intents. As suggested in~\cite{Han2019learning}, we capture such intent feature information by pre-training the model with the labeled data. Specifically, we learn the feature representations under the supervision of the cross-entropy loss. After pre-training, we remove the classifier and use the rest of the network as the feature extractor in the subsequent unsupervised clustering process.
	
	\subsubsection{Predict $K$}
	In real scenarios, we may not always know the number of new intent categories. 
	In this case, we need to determine the number of clusters $K$ before clustering. Therefore, we propose a simple and effective method to estimate $K$ with the aid of the well-initialized intent features.
	
	We assign a big $K'$ as the number of clusters (e.g., two times of the ground truth number of intent classes) at first. As a good feature initialization is helpful for partition-based methods (e.g., k-means)~\cite{platt1999probabilistic}, we use the well pre-trained model to extract intent features. Then, we perform k-means with the extracted features. We suppose that real clusters tend to be dense even with $K'$, and the size of more confident clusters is larger than some threshold $t$. Therefore, we drop the low confidence cluster which size smaller than $t$, and calculate $K$ with:
	\begin{align}
		K = \sum_{i=1}^{K'}\delta(|S_{i}| >= t),
	\end{align}
	where $|S_{i}|$ is the size of the $i^{th}$ produced cluster, and $\delta{(condition)}$ is an indicator function. It outputs 1 if $condition$ is satisfied, and outputs 0 if not. Notably, we assign the threshold $t$ as the expected cluster mean size $\frac{N}{K'}$ in this formula.
	
	%%%%%%%%%%%%table_all{results}%%%%%%%%%%%%%%%%%%%
	\begin{table*}[t!]\small
		\centering
		\begin{tabular}{@{\extracolsep{4pt}}clcccccc}
			\toprule
			\centering
			&  & \multicolumn{3}{c}{CLINC} & \multicolumn{3}{c}{BANKING}\\
			\addlinespace[0.1cm] \cline{3-5} \cline{6-8} \addlinespace[0.1cm]
			& Method & NMI  & ARI & ACC & NMI & ARI & ACC \\
			\midrule
			\multirow{7}{*}{Unsupervised.} 
			& KM  & 70.89 & 26.86 & 45.06 & 54.57 & 12.18 & 29.55 \\
			& AG  & 73.07 & 27.70 & 44.03 & 57.07 & 13.31 & 31.58 \\
			& SAE-KM & 73.13 & 29.95 & 46.75  & 63.79 & 22.85 & 38.92 \\
			& DEC  & 74.83 & 27.46 & 46.89 & 67.78 & 27.21 & 41.29 \\
			& DCN  & 75.66 & 31.15 & 49.29 & 67.54  & 26.81 & 41.99 \\
			& DAC  & 78.40 & 40.49 & 55.94 & 47.35 & 14.24 & 27.41 \\
			& DeepCluster & 65.58 & 19.11 & 35.70 & 41.77 & 8.95 & 20.69\\
			\midrule
			\multirow{6}{*}{Semi-supervised.} 
			& PCK-means  & 68.70 & 35.40 & 54.61 & 48.22 & 16.24 & 32.66\\
			& BERT-KCL  & 86.82 & 58.79 & 68.86 &  75.21 & 46.72 & 60.15 \\
			& BERT-MCL  & 87.72 & 59.92 & 69.66 & 75.68 & 47.43 & 61.14 \\
			& CDAC+  & 86.65 & 54.33 & 69.89	& 72.25	& 40.97	& 53.83 \\
			& BERT-DTC  & 90.54 & 65.02 & 74.15 & 76.55 & 44.70 & 56.51 \\
			& DeepAligned  & \textbf{93.89} & \textbf{79.75} & \textbf{86.49} & \textbf{79.56} & \textbf{53.64} & \textbf{64.90}\\
			\bottomrule
		\end{tabular}
		\caption{ \label{results-main}  
			The clustering results on two datasets. We evaluate both unsupervised and semi-supervised clustering methods.  
		}
	\end{table*}
	\begin{table*}[t!]\small
		\centering
		\begin{tabular}{@{\extracolsep{4pt}}clcccccc}
			\toprule
			\centering
			&  & \multicolumn{3}{c}{CLINC} & \multicolumn{3}{c}{BANKING}\\
			\addlinespace[0.1cm] \cline{3-5} \cline{6-8} \addlinespace[0.1cm]
			& Method & NMI & ARI & ACC & NMI & ARI & ACC \\
			\midrule
			\multirow{2}{*}{Without Pre-training} 
			& Reinitialization  & 57.80 & 9.63 & 23.02 & 34.34 & 4.49 & 13.67\\
			& Alignment & 62.53 & 14.10 & 28.63 & 36.91 & 5.23 & 15.42\\
			\midrule
			\multirow{2}{*}{With Pre-training} 
			& Reinitialization  & 82.90 & 45.67 & 55.80 & 68.12 & 31.56 & 41.32\\
			& Alignment  & \textbf{93.89} & \textbf{79.75} & \textbf{86.49} & \textbf{79.56} & \textbf{53.64} & \textbf{64.90}\\
			\bottomrule
		\end{tabular}
		\caption{ \label{results-aba-1}  
			Effectiveness of the pre-training  and the alignment strategy on two datasets. 
		}
	\end{table*}
	
	\subsection{Deep Aligned Clustering}
	After transferring knowledge from known intents, we propose an effective clustering method to find unlabeled known classes and discover novel classes. We firstly perform clustering and obtain cluster assignments and centroids. Then, we propose an original strategy to provide aligned targets for self-supervised learning.
	
	\subsubsection{Unsupervised Learning by Clustering}
	As most of the training data are unlabeled, it is important to effectively use a mass of unlabeled samples for discovering novel classes.
	
	Inspired by DeepCluster~\cite{caron2018deep}, we can benefit from the discriminative power of BERT to produce structured outputs as weak supervised signals. Specifically, we firstly extract intent features of all training data from the pre-trained model. Then, we use a standard clustering algorithm, K-Means, to learn both the optimal cluster centroid matrix $\boldsymbol{C}$ and the cluster assignments $\{y_{i}\}_{i=1}^{N}$:
	\begin{align}
		\min _{\boldsymbol{C}\in \mathds R^{K \times D}}\frac{1}{N}\sum_{i=1}^{N}\min _{y_{i}\in\{1, \ldots, K\}}\left\|\boldsymbol{I}_i-\boldsymbol{C}_{y_{i}}\right\|_{2}^{2},
		\label{k-means}
	\end{align}
	where $N$ is the number of training samples and  $\|\cdot\|_{2}^{2}$ denotes the squared Euclidean distance. Then, we leverage the cluster assignments as pseudo-labels for feature learning. 
	
	\subsubsection{Self-supervised Learning with Aligned Pseudo-labels}
	DeepCluster alternates between clustering and updating network parameters. It performs k-means to produce cluster assignments as pseudo-labels and uses them to train the neural network. However, the indices after k-means are permuted randomly in each training epoch, so the classifier parameters have to be reinitialized before each training epoch~\cite{Zhan_2020_CVPR}. Thus, we propose an alignment strategy to tackle the assignment inconsistency problem.
	
	We notice that DeepCluster lacks the use of the centroid matrix $\boldsymbol{C}$ in Eq.~\ref{k-means}. However, $\boldsymbol{C}$ is a crucial part, which contains the optimal averaged assignment target of clustering. As each embedded sample is assigned to its nearest centroid in Euclidean space, we naturally adopt $\boldsymbol{C}$ as the prior knowledge to adjust the inconsistent cluster assignments in different training epochs. That is, we convert this problem into the centroid alignment. Though the intent representations are updated continually, similar intents are distributed in near locations. The centroid synthesizes all similar intent samples in its cluster, so it is more stable and suitable for guiding the alignment process. We suppose the centroids in contiguous training epochs are relatively consistently distributed in Euclidean space, and adopt the Hungarian algorithm~\cite{kuhn1955hungarian} to obtain the optimal mapping $G$:
	\begin{align}
		\boldsymbol{C}^{c} = G(\boldsymbol{C}^{l}),
	\end{align}
	where $\boldsymbol{C}^{c}$ and $\boldsymbol{C}^{l}$ respectively denote the centroid matrix in the current and last training epoch. Then, we obtain the aligned pseudo-labels $y^{align}$ with $G(\cdot)$:
	\begin{align}
		y^{align} = G^{-1}(y^{c}),
	\end{align}
	where $G^{-1}$ denotes the inverse mapping of $G$ and $y^{c}$ denotes the pseudo-labels in the current training epoch. Finally, we use the aligned pseudo-labels to perform self-supervised learning under the supervision of the softmax loss $\mathcal{L}_{s}$:
	\begin{align}
		\mathcal{L}_{s}=-\frac{1}{N}\sum_{i=1}^{N} \log\frac{\exp(\phi(\boldsymbol{I}_{i})^{y_{i}^{align}})}{\sum_{j=1}^{K}\exp(\phi(\boldsymbol{I}_{i})^{j})},
	\end{align}
	where $\phi(\cdot)$ is the pseudo-classifier for self-supervised learning, and $\phi(\cdot)^{j}$ denotes the output logits of the $j^{th}$ class.
	
	We use the cluster validity index (CVI) to evaluate the quality of clusters obtained during each training epoch after clustering. Specifically, we adopt an unsupervised metric Silhouette Coefficient~\cite{ROUSSEEUW198753} for evaluation:
	\begin{align}
		SC=\frac{1}{N}\sum_{i=1}^{N} \frac{b(\boldsymbol{I}_{i})-a(\boldsymbol{I}_{i})}{\max \{a(\boldsymbol{I}_{i}), b(\boldsymbol{I}_{i})\}},
		\label{8}
	\end{align}
	where $a(\boldsymbol{I}_{i})$ is the average distance between $\boldsymbol{I}_{i}$ and all other samples in the $i^{th}$ cluster, which indicates the intra-class compactness. $b(\boldsymbol{I}_{i})$ is the smallest distance between $\boldsymbol{I}_{i}$ and all samples not in the $i^{th}$ cluster, which indicates the inter-class separation. The range of $SC$ is between -1 and 1, and the higher score means the better clustering results.

	\section{Experiments}
	
	\subsection{Datasets}
	We conduct experiments on two challenging benchmark intent datasets. Detailed statistics are shown in Table ~\ref{datasets}.
	\subsubsection{CLINC} It is an intent classification dataset~\cite{larson-etal-2019-evaluation}, which contains 22,500 queries covering 150 intents across 10 domains.
	\subsubsection{BANKING} It is a fine-grained dataset in the banking domain~\cite{Casanueva2020}, which contains 13,083 customer service queries with 77 intents.
	\subsection{Baselines}
	\subsubsection{Unsupervised}
	We firstly compare with unsupervised clustering methods, including K-means (KM)~\cite{macqueen1967some}, agglomerative clustering (AG)~\cite{gowda1978agglomerative}, SAE-KM, DEC~\cite{xie2016unsupervised}, DCN~\cite{yang2017towards}, DAC~\cite{chang2017deep}, and DeepCluster~\cite{caron2018deep}.

	\begin{figure*}[t!]
		\centering 
		\includegraphics[scale=.28]{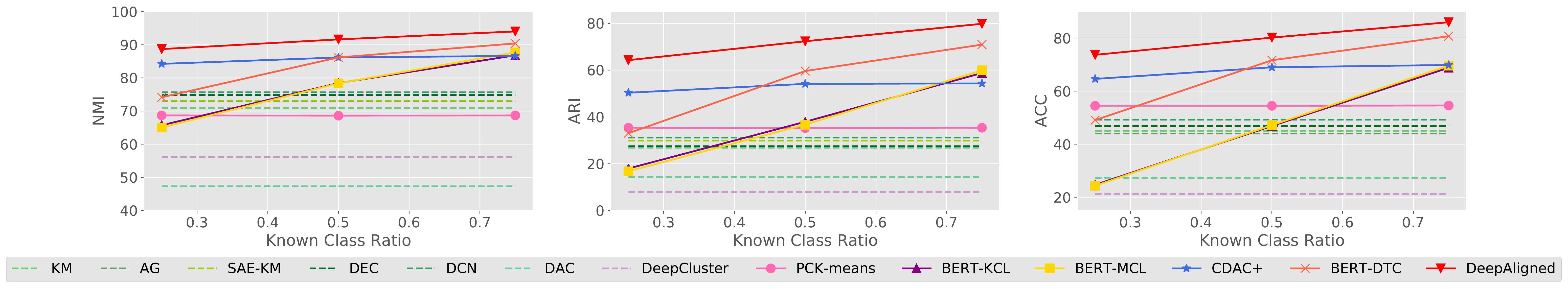}
		\caption{\label{results-aba-2-1} Influence of the known class ratio on CLINC dataset.}
	\end{figure*}
	\begin{figure*}[t!]
		\centering  
		\includegraphics[scale=.28]{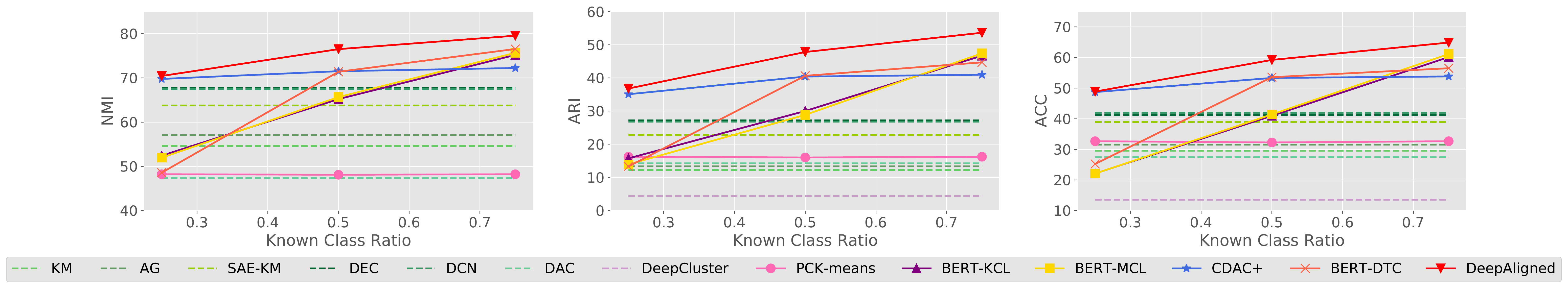}
		\caption{\label{results-aba-2-2} Influence of the known class ratio on BANKING dataset.}
	\end{figure*}
	
	For KM and AG, we represent the sentences with the averaged pre-trained 300-dimensional word embeddings from GloVe~\cite{pennington2014glove}. For SAE-KM, DEC, and DCN, we encode the sentences with the stacked autoencoder (SAE), which is helpful to capture meaningful semantics on real-world datasets~\cite{xie2016unsupervised}. As DAC and DeepCluster are unsupervised clustering methods in computer vision, we replace the backbone with the BERT model for extracting text features.
	\begin{table}\small
		\centering
		\begin{tabular}{@{\extracolsep{0.6pt}}clcccc}
			\toprule
			\centering
			% &  & \multicolumn{2}{c}{Full} & \multicolumn{2}{c}{Small} & \multicolumn{2}{c}{Imbal} & \multicolumn{2}{c}{OOS+} \\
			&  & \multicolumn{2}{c}{CLINC (K'=300)} & \multicolumn{2}{c}{BANKING (K'=154)} \\
			\addlinespace[0.1cm] \cline{3-4} \cline{5-6}  \addlinespace[0.1cm]
			& Methods &  K (Pred) & Error& K (Pred) & Error\\
			\midrule
			\multirow{3.5}{*}[1ex]{\rotatebox[origin=c]{90}{\it 25\%}} 
			%		\multirow{3.5}{*}[2ex]{\it 25\%}
			&  BERT-MCL & 38 & 75.00 & 19 & 75.32 \\
			&  BERT-DTC & 94 & 37.33 & 37 & 51.95 \\							
			& DeepAligned  & \textbf{122} & \textbf{18.67} & \textbf{66} & \textbf{14.29}\\
			\midrule
			\midrule
			\multirow{1.5}{*}[-1.5ex]{\rotatebox[origin=c]{90}{\it 50\%}}
			& BERT-MCL & 75	& 50.00	& 38 & 50.65	\\
			&  BERT-DTC & \textbf{131} & \textbf{12.67} & \textbf{71} & \textbf{7.79} \\							
			& DeepAligned &130	&13.33	&64	&16.88\\
			\midrule
			\midrule
			\multirow{1.5}{*}[-1.5ex]{\rotatebox[origin=c]{90}{\it 75\%}}
			& BERT-MCL & 112 & 25.33 	& 58 &  24.68 \\
			&  BERT-DTC & 195 & 30.00 & 110 & 42.86 \\							
			& DeepAligned &\textbf{129}	&\textbf{14.00}	&\textbf{67}	&\textbf{12.99}	\\
			\bottomrule
		\end{tabular}
		\caption{  
			The results of predicting $K$ with an unknown number of clusters. We vary the known class ratio in the range of 25\%, 50\% and 75\%, and set $K'$ as two times of the ground truth number of clusters during clustering. 
		}
		\label{aba-4} 
	\end{table}
	\subsubsection{Semi-supervised}
	We also compare our method with semi-supervised clustering methods, including PCK-means~\cite{basu2004active}, BERT-KCL~\cite{hsu2018learning}, BERT-MCL~\cite{hsu2018multiclass}, BERT-DTC~\cite{Han2019learning} and  CDAC+~\cite{lin2020discovering}. For a fairness comparison, we replace the backbone of these methods with the same BERT model as ours.
	
	\subsection{Evaluation Metrics}
	We adopt three widely used metrics to evaluate the clustering results: Normalized Mutual Information (NMI), Adjusted Rand Index (ARI), and Accuracy (ACC). To calculate ACC, we use the Hungarian algorithm to obtain the mapping between the predicted classes and ground-truth classes.
	
	\subsection{Evaluation Settings}
	Following the same settings as in~\cite{lin2020discovering}, we randomly select 10\% of training data as labeled and choose 75\% of all intents as known. We split datasets into the training, validation, and test sets. The number of intent categories is set as ground-truth. We first use the little labeled data of known intents for pre-training, and tune with the validation set. Then, we use all training data for self-supervised learning and evaluate the cluster performance with Silhouette Coefficient (as mentioned in Eq.~\ref{8}). Finally, we evaluate the performance on the test set and report the averaged results over ten runs of experiments with different random seeds.
	\subsection{Implementation Details}
	We use the pre-trained BERT model (bert-uncased, with 12-layer transformer) implemented in PyTorch~\cite{Wolf2019HuggingFacesTS} as our network backbone, and adopt most of its suggested hyper-parameters for optimization. The training batch size is 128, the learning rate is $5e^{-5}$, and the dimension of intent features $D$ is 768. Moreover, as suggested in~\cite{lin2020discovering},  we freeze all but the last transformer layer parameters to speed up the training procedure and improve the training efficiency with the backbone of BERT.

	\section{Results and Discussion}
	Table~\ref{results-main} shows the results of all compared methods. We highlight the best results in bold. Compared with baselines, our method consistently achieves the best results and outperforms other baselines by a large margin on all metrics and datasets. It demonstrates the effectiveness of our method to discover new intents with limited known intent data. We also find most semi-supervised methods perform better than unsupervised methods. It indicates that even with limited labeled data as prior knowledge, it is also helpful to improve the performance of unsupervised clustering.
	
	\begin{figure*}[t!]
		\centering 
		\includegraphics[scale=.28]{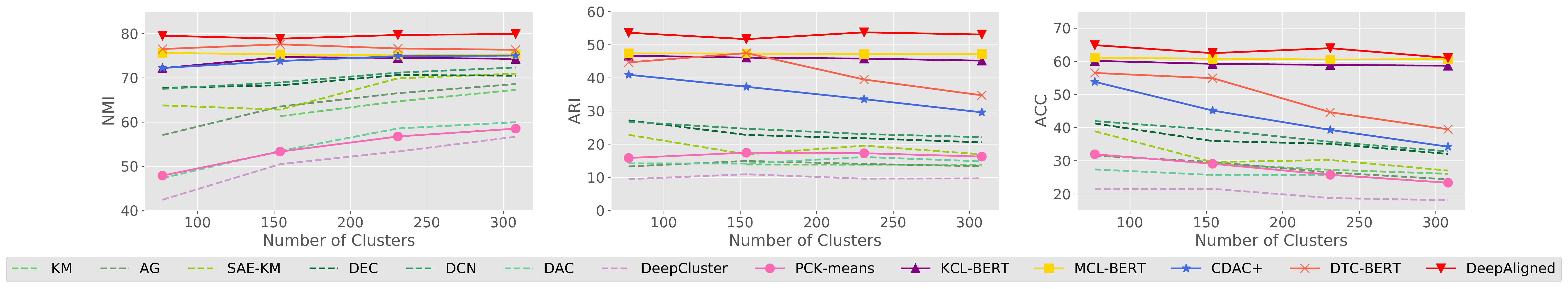}
		\caption{\label{results-aba-3-1} Influence of the number of clusters on BANKING dataset.}
	\end{figure*}
	\begin{figure*}[t!]
		\centering  
		\includegraphics[scale=.282]{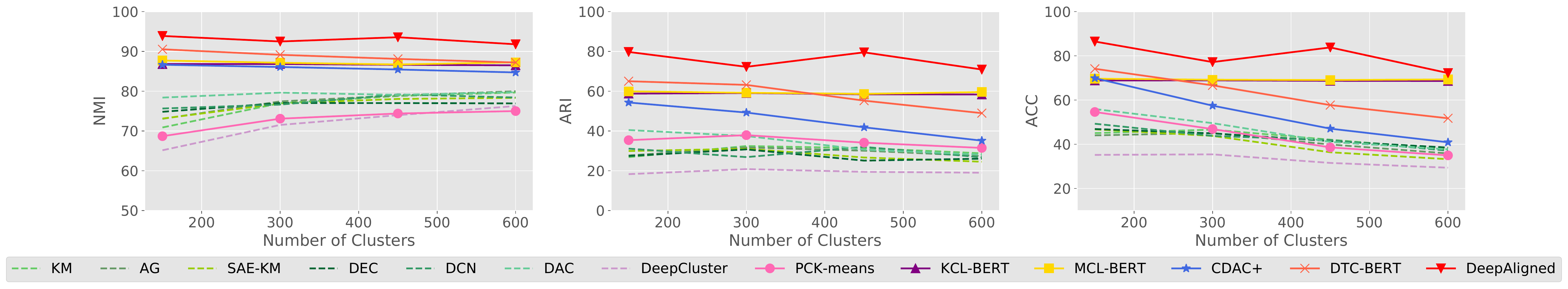}
		\caption{\label{results-aba-3-2} Influence of the number of clusters on CLINC dataset.}
	\end{figure*}
	
	\subsection{Effect of the Alignment Strategy}
	To investigate the contribution of the alignment strategy, we compare our method with the reinitialization strategy~\cite{caron2018deep}. As shown in Table~\ref{results-aba-1}, our method has significant improvements over the reinitialization strategy on both semi-supervised and unsupervised settings. We suppose the reason is that random initialization drops out the well-trained parameters in the classifier in the former epochs. By contrast, our method saves the history embedding information by finding the mapping of produced pseudo-labels between contiguous epochs, which provides stronger supervised signals for representation learning. 
	
	\subsection{Estimate $K$} 
	To investigate the effectiveness to predict $K$, we assign $K'$ as two times of the ground truth number of intent classes and compare with another two state-of-the-art methods (BERT-MCL and BERT-DTC). We vary the ratio of known classes in the range of 25\%, 50\%, and  75\%, and calculate the error rate (the lower is better) for evaluation. As shown in Table~\ref{aba-4}, our method achieves the alomost the lowest error rates with different known class ratios. It shows the reasonability to estimate the cluster number by removing low-confidence clusters with well-initialized intent features. We notice that BERT-DTC is a strong baseline, especially with 50\% known classes. The reason is that BERT-DTC also relies the labeled samples to generate the probe set for determining the optimal number of classes. Nevertheless, the performance is unstable. We also find the predicted $K$ of BERT-MCL is close to the number of known classes. The reason is that BERT-MCL jointly performs clustering and classification.  However, the classification part dominates in training under the supervision of labeled data, so it tends to misclassifies new intents into known intents during testing.
	
	\subsection{Effect of the Known Class Ratio}
	To investigate the influence of the number of known intents, we vary the known class ratio in the range of 25\%, 50\% and 75\% during training. As shown in Figure~\ref{results-aba-2-1} and Figure~\ref{results-aba-2-2}, our method achieves the best results with different number of known intents. All semi-supervised methods are sensitive to the number of known intents. Particularly, though BERT-MCL and BERT-DTC achieve competitive results with 75\% known intents, their performances drop dramatically as the known class ratio decreases. We suppose the reason is that they largely depend on the prior knowledge of known intents to construct supervised signals (e.g., the pairwise similarities in BERT-MCL and the initialized centroids in BERT-DTC) for clustering. Therefore, the learned features of these methods are much more biased towards the labeled data.
	
	By contrast, our method only needs labeled  intent data for learning feature representations. Thus, it is free from the bias towards labeled data during self-supervised learning process. Moreover, our method achieves more robust results with fewer known intents.
	
	\subsection{Effect of the Number of Clusters}
	To investigate the sensitiveness of the assigned cluster number $K'$, we vary $K'$ from the ground-truth number to four times of it. The known class ratio is assigned as 75\%. As shown in Figure~\ref{results-aba-3-1} and Figure~\ref{results-aba-3-2}, our method achieves the best results with different number of assigned clusters. We notice that most semi-supervised clustering methods are vulnerable to the number of clusters, and their performances drop to some extent with large  $K'$. It is because many redundant classes may result in splitting fine-grained clusters of originally one cluster with the same intent-label. Compared with all these methods, our method benefits from a more accurate estimated cluster number for clustering. Therefore, it achieves better results even with a large  $K'$.

	\section{Conclusion and Future Work}
	In this work, we have introduced an effective method for discovering new intents. Our method successfully transfers the prior knowledge of limited known intents and estimates the number of intents by eliminating low-confidence clusters. Moreover, it provides more stable and concrete supervised signals to guide the clustering process. We conduct extensive experiments on two challenging benchmark datasets to evaluate the performance. Our method achieves significant improvements over the compared methods and obtains more accurate estimated cluster numbers with limited prior knowledge. In the future, we will try different clustering methods to produce supervised signals and explore more self-supervised methods for representation learning.

	\section{Acknowledgments}
	This work is supported by seed fund of Tsinghua University (Department of Computer Science and Technology)-Siemens Ltd., China Joint Research Center for Industrial Intelligence and Internet of Things. 
	\bibliography{aaai21.bib}
\end{document}